\documentclass[letterpaper, 10 pt, conference]{ieeeconf}

\IEEEoverridecommandlockouts
\usepackage{comment}
\usepackage{gensymb}
\usepackage{graphics}
\usepackage{epsfig}
\usepackage{mathptmx}
\usepackage{times}
\usepackage{amsmath}
\usepackage{amssymb}
\usepackage{xcolor}
\usepackage{subcaption}
\usepackage{graphicx}
\usepackage{wrapfig}
\usepackage{caption}
\usepackage{soul}
\usepackage{hyperref}
\usepackage[title]{appendix}

\DeclareMathAlphabet{\mathcal}{OMS}{cmsy}{m}{n}

\title{\LARGE \bf
Few-Shot Keypoint Detection as Task Adaptation via Latent Embeddings
}

\newcommand{\ie}{\textit{i}.\textit{e}. }
\newcommand{\eg}{\textit{e}.\textit{g}. }
\newcommand\sref{Section~\ref}
\newcommand\eref{Eq.~\ref}
\newcommand\fref{Fig.~\ref}

\newcommand{\Tau}{\mathcal{T}}
\newcommand{\img}{\mathcal{I}}

\author{Mel Vecerik$^{1,2}$ and Jackie Kay$^{1,2}$ and Raia Hadsell$^{2}$ and Lourdes Agapito$^{1}$ and Jon Scholz$^{2}$
\thanks{$^{1}$University College London, UK $^{2}$DeepMind, London, UK}%
\thanks{*\small Correspondence to: \{vec,jscholz\}@google.com}}

\begin{document}

\maketitle
\thispagestyle{empty}
\pagestyle{empty}

\begin{abstract}

Dense object tracking, the ability to localize specific object points with pixel-level accuracy, is an important computer vision task with numerous downstream applications in robotics.
Existing approaches either compute dense keypoint embeddings in a single forward pass, meaning the model is trained to track everything at once, or allocate their full capacity to a sparse predefined set of points, trading generality for accuracy.
In this paper we explore a middle ground based on the observation that the number of relevant points \textit{at a given time} are typically relatively few, \eg grasp points on a target object.
Our main contribution is a novel architecture, inspired by few-shot task adaptation, which allows a sparse-style network to condition on a keypoint embedding that indicates which point to track.
Our central finding is that this approach provides the generality of dense-embedding models, while offering accuracy significantly closer to sparse-keypoint approaches.
We present results illustrating this capacity vs. accuracy trade-off, and demonstrate the ability to zero-shot transfer to new object instances (within-class) using a real-robot pick-and-place task.
\end{abstract}

\section{Introduction}
\label{sec:introduction}

\begin{wrapfigure}{t}{2.5cm}
    \vspace{-0.6cm}
    \includegraphics[width=\linewidth]{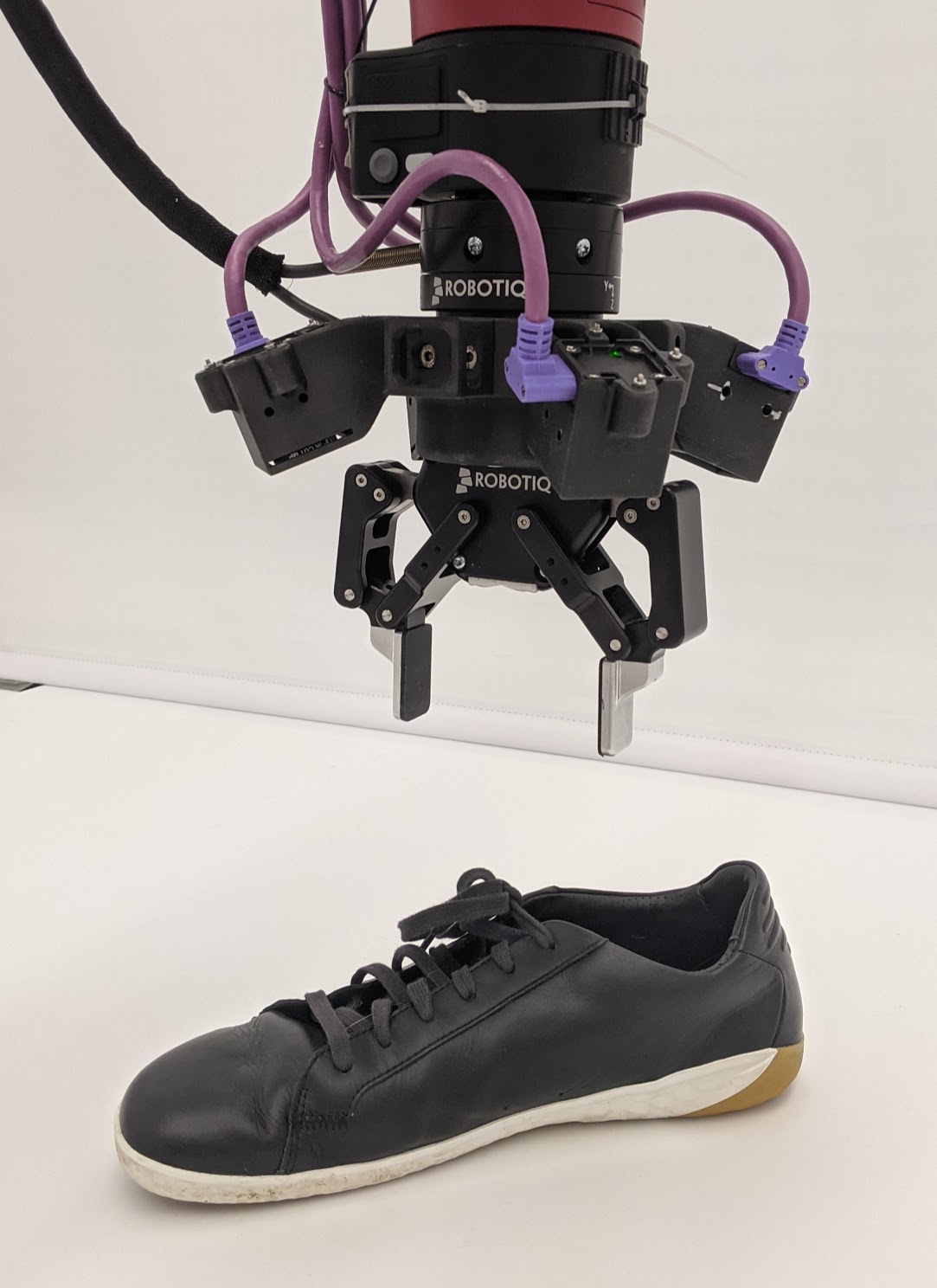}
    \caption{Robot grasping setup.}
    \label{fig:robot_picture}
    \vspace{-0.2cm}
\end{wrapfigure}

When teaching another human a new manual task, we can succinctly describe the task-relevant object parts, because of our shared implicit understanding of the 3D world.
Yet in robotics, we define complex perception pipelines to extract the relevant 3D locations on object parts, \eg cable sockets, or corners of a piece of wood.
There is currently no method which can adapt to detect such keypoints within a few annotations in a spatially consistent, three-dimensional manner that is robust to occlusions.
Rapid adaptation to new points is critical for fast iteration when defining robot motions, therefore few-shot task adaptation techniques are a natural candidate for this problem. 

There are a variety of relevant approaches to this problem: predicting the poses of rigid constituents~\cite{byravan2017se3}, reasoning about objects on a per-pixel basis in the camera view \cite{florencemanuelli2018dense, jabri2020spacetime}, or tracking a fixed subset of points \cite{Manuelli2019kPAMKA, NIPS2018_7476, superpoint18_detone}.
However object poses are well-defined only for rigid objects, per-pixel representations struggle with occlusions, and tracking specific points does not generalize well to novel points.
At the same time, for a representation to be actionable by a robot end effector controller, it must be three-dimensional and must also generalize to novel viewpoints and objects.will have beef with this

\begin{figure}[t]
  \vspace{-0.2cm}
  \centering
\includegraphics[width=0.48\textwidth]{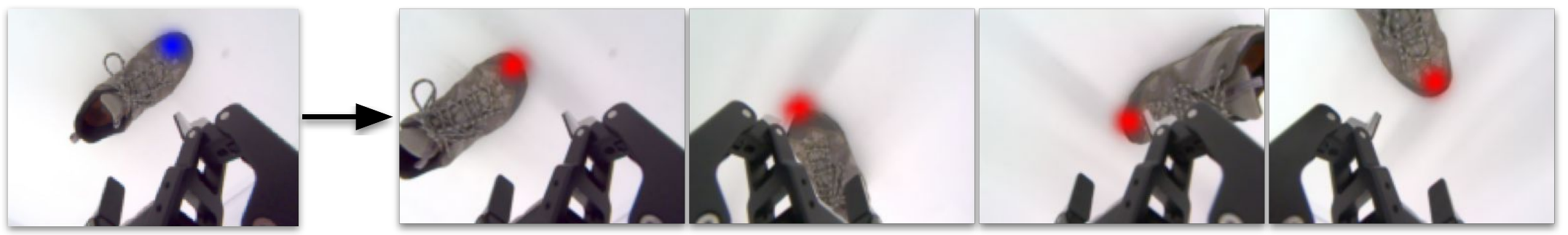}
    \caption{
        Using TACK we extract an embedding from a single annotation (blue) and detect corresponding points (red) in novel views even when large parts of the object are occluded.
        This model was trained on a combination of synthetic and real data, but has never seen this shoe before.
    }
  \label{fig:real_shoe}
  \vspace{-0.6cm}
\end{figure}

In this work, we formulate keypoint tracking as a task adaptation problem where each task corresponds to inferring the 3D location of a specific keypoint from any view.
This can be also be seen as meta-learning where the meta-adaptation happens via a latent space variable \cite{pmlr-v97-rakelly19a}.
We split this problem into two stages where first we infer the identity of the point which needs to be detected and then detect it from novel views.
These stages are trained end to end via a combination of task adaptation and conditional autoencoder losses.
We show properties of this model and demonstrate its viability on a robot grasping task (\fref{fig:robot_picture}) where we use the ability to detect non-surface points on novel object instances based only on a single annotation, visualized in \fref{fig:real_shoe}.
We call this approach Task Adaptation for Conditioned Keypoints, or TACK.
Videos and further materials available at \href{https://sites.google.com/view/2021-tack/home}{\color{blue}{sites.google.com/view/2021-tack}}.

\section{Related work}
\label{sec:related_work}

Tracking keypoints for robotics has been previously explored via several approaches.
kPAM \cite{Manuelli2019kPAMKA} demonstrated learning keypoints defined via full supervision by human annotation for scripting robot grasping controllers.
S3K \cite{vecerik2020s3k} approached a similar problem by using geometric self-supervision to minimize the number of required annotations.
This approach demonstrated that only ten 3D-labelled samples are enough to semantically ground the model and allow further improvement using purely unsupervised data, enabling a robot to solve tasks such as cable insertion or robot grasping.
However providing 10 labelled samples per keypoint in addition to an unsupervised corpus of real robot data is too time-consuming to allow for fast iteration as it requires full retraining if different keypoints are to be tracked.
Although kPAM and S3K both empirically test generalization to unseen objects, neither method provides an explicit mechanism for transferring knowledge to track new points on the object or a way to generalize across instances without explicit labels.

Another family of approaches is dense descriptor methods, where an embedding is defined for every pixel in the image.
The Dense Object Nets~\cite{florencemanuelli2018dense} utilized two loss functions: one to ensure that the same point from different views has the same embedding and another which ensured that different points had different embeddings.
Keypoint detection is accomplished by retrieving the embedding of a pixel of interest and locating the pixel with the most similar embedding in a novel view, enabling point localization using only a single annotation.
This method does can not track points which are not on the visible surface, \eg the inside of a socket in a plug insertion task \cite{Vecerk2019APA}.
Other approaches learn dense embeddings via self-supervision and large amounts of video data \cite{jabri2020spacetime}, however they usually do not leverage the 3D geometry of the task which, as has been shown in S3K\cite{vecerik2020s3k} can greatly improve sample efficiency.

Other works have applied few-shot learning methods, in particular meta-learning, to track objects from vision.
For example, in \cite{wang2020tracking} an object detector model was adapted in several steps via gradient-based meta-learning to allow tracking of a novel object.
However, this method
does not provide a good way to leverage the known 3D structure of the object, which in our method is accomplished via a low-dimensional latent space.
Our training method draws inspiration from few-shot learning methods via generative modelling.
\cite{Gordon19metalearning} is a similar setting to TACK: a handful of images are embedded into a latent variable, and the decoder infers a generated image from it. However TACK solves a detection problem, rather than image generation.
A similar problem is addressed in \cite{sohn2015cvae}, where a variational autoencoder is conditioned on additional information to guide the latent space's structure.
Our approach separates the identity and location of keypoints and encourages the latent space to focus only on identity using similar conditioning.

\section{Method}

The goal of TACK is to learn to track new keypoints given a few annotations of the same point from different viewpoints.
We cast dense keypoint tracking as a few-shot task adaptation
and consider a distribution of tasks $\rho(\Tau)$ where each task $\Tau$ corresponds to a 3D location.
Our overall objective is to be able to adapt to a novel locations and instances based only on a few observations.

\subsection{Problem Setting}
\label{sec:dataset}

To learn to adapt to new tasks our training dataset has to contain both \emph{training}
and a \emph{validation} set for each task.
The task adaptation loss is based on the model's ability to generalize to the validation set given the task conditioning instances.
In addition to this we include a conditioned autoencoder loss which is based purely on the task training set.
The observation for task $\Tau$ is a batch of images $\img^\Tau_i$ and targets $t^\Tau_i$ for $i \in [1, L]$ where $L$ is the number of input views.
We also have an evaluation dataset with a similar training and validation split which uses a different set of objects.

\subsection{Dataset Generation}
\label{sec:simulation_dataset}

We build our dataset using the shoe category of the Google Scanned Objects dataset \cite{IgnitionFuel-GoogleResearch-Google-Scanned-Objects}, which contains 56 scanned shoes for training and 8 for evaluation.
To generate a sample, we instantiate a task $\Tau$ by sampling a shoe instance and a single point $x^\Tau$ on its surface. 
We then randomize the shoe position within a bounding box with limits chosen such that the shoe's size is between 20-60\% of the image width when rendered.
The shoe's orientations are correlated within the task to ensure that similar parts are visible.
This is done by first sampling a quaternion $q_\tau$, then drawing a sample from $\mathcal{N}(q_\tau, 0.2^2)$, and normalizing the result.
Afterwards we perform a random in-plane rotation which preserves the visible parts of the object.
We only use images where the keypoint's projected position lies within the FOV of the camera.
Note that this doesn't mean that the point has to be visible - it can be on the far side of the object.
The result of rendering this scene is a 2D image $\img^\Tau$.

After selecting the 3D pose of the shoe in the scene and the 3D point of interest on the surface $x^\Tau$, we project the point onto a 2D target heatmap. 
Using camera calibration we define the projection points to have coordinates $i^x$ and $j^x$ which allows us to define our target image of the same spatial shape $\img^i$ as 
$t_{ij} = \exp{\left( ((i - i_x)^2 + (j - j_x)^2)/(2 \sigma ^ 2) \right)}$.
At each sampling step, we produce 4 frames $\img^\Tau_i$ and 4 corresponding point projection images $t^\Tau_i$.
$L=3$ of these pairs form the train dataset for task training and 1 is used as the task validation, \ie the target for few-shot adaptation.
We use an image resolution of 160x120 and our training dataset contains 1M of these batches.
Additionally, we pad the sampled image by 8 pixels on each side and randomly crop it to recover original resolution, which reduces over-fitting \cite{yarats2021image}.
Additionally we also produce a segmentation mask and predict it for each pixel as an auxiliary task for the decoder, which helps to stabilize learning.

\begin{figure}[]
  \centering
  \vspace{-0.1cm}
\includegraphics[width=0.45\textwidth]{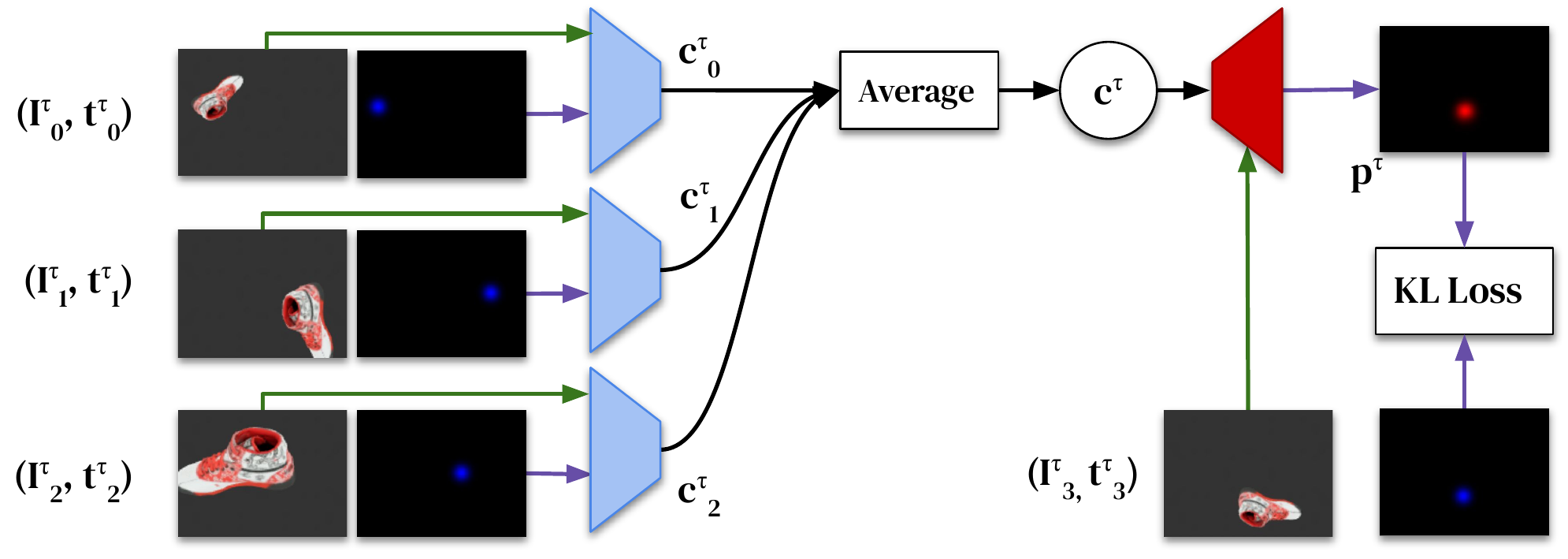}
  \vspace{-0.2cm}
    \caption{
    Task adaptation setting for TACK. 
    Image, target pairs are separately encoded and task embeddings are averaged. 
    The decoder is conditioned on an image from $\mathcal{D}^\Tau_{valid}$, and the KL divergence between the validation target and the decoded prediction is minimized.
    }
  \label{fig:task_adaptation}
  \vspace{-0.6cm}
\end{figure}

\subsection{Network Architectures}
\label{sec:network_architecture}

The task of detecting a keypoint in a novel image can be split into two sub-problems: understanding which point to track and localizing it in a novel view.
We separate these responsibilities into two modules: a task encoder $Enc$ which encodes the task into a latent variable and a decoder $Dec$ which uses the task latent variable to detect the keypoint in a given image.

The encoder network combines information from two images, object image $\img^\Tau$ and projection $t^\Tau$, and generates the task embedding $c^\Tau$.
Since these images are spatially aligned, a natural choice is to concatenate them in the channel dimension and encode them using a ResNet \cite{he2016deep}. 
The first layer projects the input to 32 channels and each subsequent layer applies a residual block, ReLu nonlinearity, halves the spatial resolution, and doubles the number of channels.
The last layer is a spatial max and an MLP to decrease the number of channels to required embedding size \cite{finn2016deep}.

The decoder network accepts the image $\img^\Tau$ in which the point should be detected and the encoding $c^\Tau$ which defines which point should be tracked.
The output of the decoder is a single channel image $p^\Tau$ containing a heatmap for the predicted projection of the point $x^\Tau$.
The structure of this problem is closely related to image segmentation, therefore we use a Residual U-net as our architecture \cite{zhang18unet}.
We have found that putting a FiLM layer \cite{Perez2018film} before every residual block gave us the best results, similar to \cite{stylegan2019karras, Burkov_2020_CVPR}.
This architecture is similar to the encoder except it also includes a second structurally symmetrical half for up-sampling the resolution, with skip connections between blocks.
Final keypoint position is inferred via soft argmax.

\subsection{Task Adaptation}
\label{sec:task_adaptation}

This problem is split into inferring the task latent code $c^\Tau \in \mathbb{R}^k$ from annotations and using it to detect the keypoint from a novel view \fref{fig:task_adaptation}.
First, the encoder receives pairs consisting of the camera image and a heatmap image.
The task latent embedding $c^\Tau$ representing the task $\Tau$ is an average of the $L$ individual encoder outputs for each input pair drawn from the training distribution.
The decoder has to solve the task $\Tau$ by detecting the point $x^\Tau$ when conditioned by the appropriate task embedding vector $c^\Tau$.
Therefore we use the decoder to give us prediction $p^\Tau$ for the projection location in a specific image $\img^\Tau$ which allows us to define a loss between the prediction and the target $t^\Tau$ corresponding to $\img^\Tau$ as:

\vspace{-0.7cm}
\begin{align}
\label{eq:adaptation_loss}
    c^\Tau = \sum_i^L Enc(\img^\Tau_i, t^\Tau_i) / L \nonumber \\
    \mathcal{L}_{adapt} = D_\mathrm{KL}(t^\Tau_{L+1} | Dec(c^\Tau, \img^\Tau_{L+1}))
\end{align}

\subsection{Conditional Autoencoder}
\label{sec:conditional_autoencoder}

\begin{figure}[]
  \centering
  \vspace{0.2cm}
\includegraphics[width=0.5\textwidth]{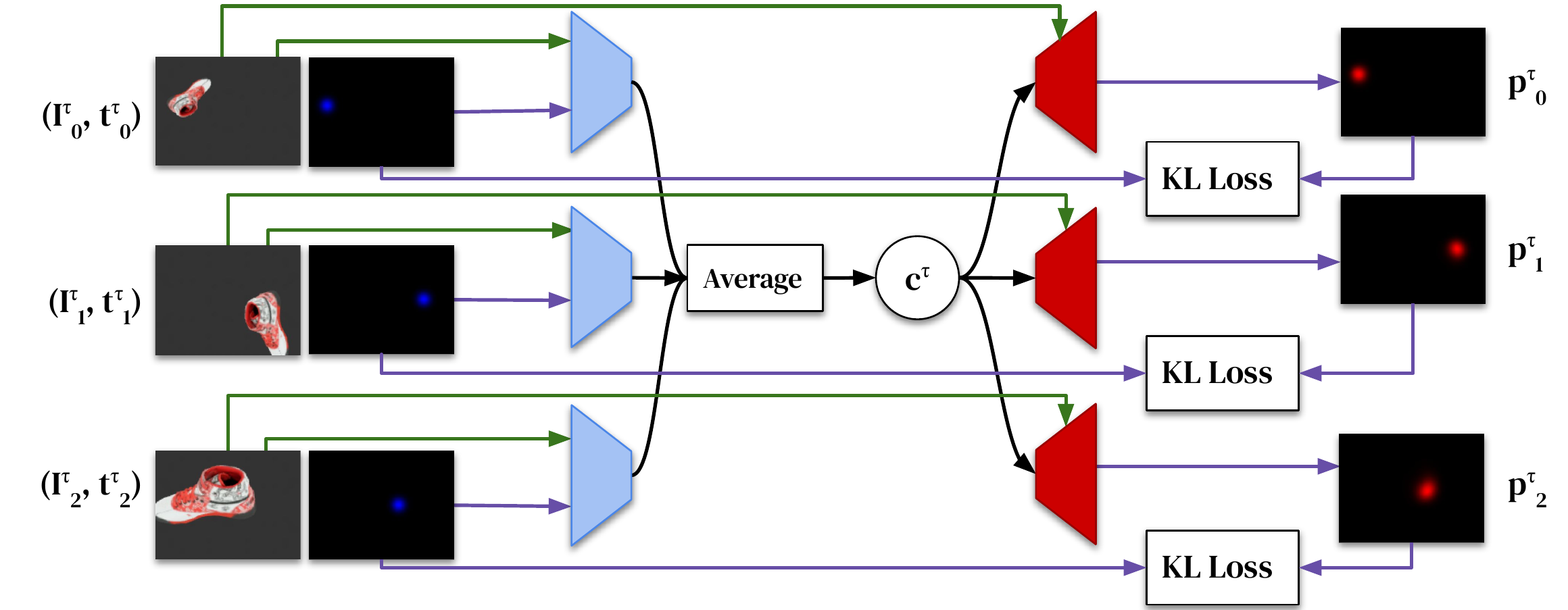}
    \caption{
        Conditional autoencoder setting.
        (Image, target) pairs are encoded separately and embeddings are averaged.
        The decoder uses the combined embedding while separately conditioning on each input image.
        The KL loss for each pair of input targets and decoded predictions is minimized.
    }
  \label{fig:conditional_autoencoder}
  \vspace{-0.5cm}
\end{figure}

We have explained how this problem can be formulated as task adaptation.
However, task adaptation methods can be difficult to train due to noisy gradients.
An alternative formulation that may be easier to train, but does not capture the few-shot learning objective, is to directly condition the decoder on the inputs.
That is, we condition the decoder with the same image used as a task conditioning input to the encoder.
Since we are still considering an average embedding from a batch of image-target pairs, we decode the combined $c^\Tau$ multiple times with different conditioning and sum the $\mathcal{D}_{KL}$ losses given the respective target, as follows:

\vspace{-0.4cm}
\begin{align}
\label{eq:autoencoder_loss}
    c^\Tau = \sum_i^L Enc(\img^\Tau_i, t^\Tau_i) / L  \nonumber \\
    \mathcal{L}_{auto-enc} = \sum_i^L D_{\mathrm{KL}} \left ( t^\Tau_i | Dec(c^\Tau, \img^\Tau_i) \right)
\end{align}
\vspace{-0.4cm}

This is recognizable as a conditional autoencoder, as seen in \fref{fig:conditional_autoencoder}.
From this perspective the embedding has to learn about the semantic identity of the point, because it has to be consistent across the batch.
However, it should not have to contain information about the specific projection locations within the images, as these can be inferred within the decoder from the conditioning images $\img^\Tau_i$.
In practice we always use a combination of both \eref{eq:adaptation_loss} and \eref{eq:autoencoder_loss} as discussed in \sref{sec:meta_vs_autoencoder}.

\vspace{-0.1cm}

\section{Experiments}
\label{sec:experiments}

\vspace{-0.1cm}

To present TACK we performed a number of experiments on synthetic datasets and demonstrated its performance in a real robot setting.
The aim of these experiments is to provide intuition for properties of the model such as its accuracy under various conditions and the structure of the latent space which is learned.

For the number of task conditioning annotations $L$ drawn from the training distribution, we found that $L=3$ was sufficient for training and evaluation.
However we found that $L=1$ can be enough to identify the point to track.
We chose the RMS pixel error between the soft-argmax of the prediction and ground truth as our evaluation metric, because it can be evaluated on a per-image bases without the need for 3D geometry.
Since regardless of the loss function the model always outputs a heatmap of the target keypoint, the pixel-wise RMSE allows us to directly compare ablations of the model.

\vspace{-0.1cm}

\subsection{Task Adaptation vs. Conditional Autoencoder}
\label{sec:meta_vs_autoencoder}

\begin{figure*}[]
  \centering
    \vspace{0.2cm}
\includegraphics[width=0.9\textwidth]{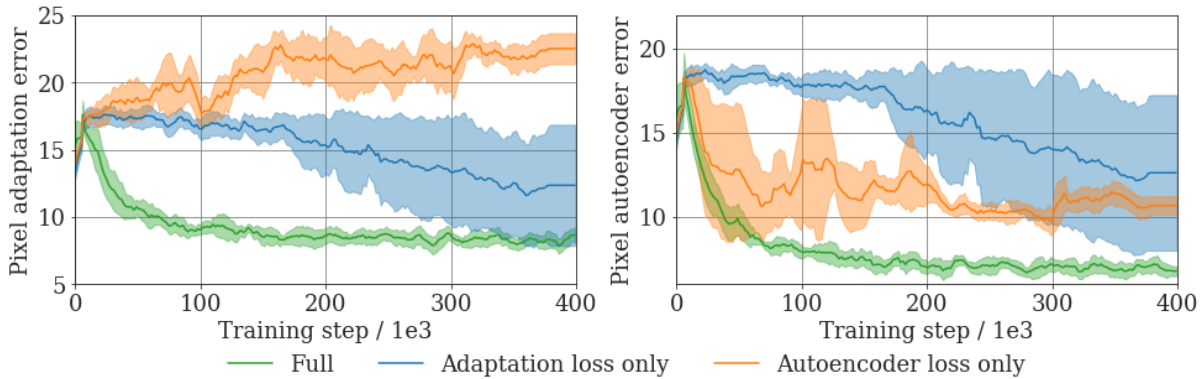}
    \vspace{-0.2cm}
    \caption{
        RMS pixel error for 3 different training regimes.
        Left: Evaluated in the task adaptation setting as in \sref{sec:task_adaptation}. 
        Right: Evaluated in the conditional autoencoder setting as in \sref{sec:conditional_autoencoder}. 
        In both cases, training with a combination of the adaptation and autoencoder losses is beneficial.
    }
  \label{fig:meta_vs_autoencoder}
  \vspace{-0.5cm}
\end{figure*}

In \sref{sec:task_adaptation} and \sref{sec:conditional_autoencoder}, we derived a task adaptation loss and a conditional autoencoder loss for this model and described potential advantages and disadvantages of these approaches. Our hypothesis is that a combination of these losses leads to stable, fast training.

In \fref{fig:meta_vs_autoencoder} we show evaluation curves during training comparing the combined loss with ablated models that use only one of the losses.
As these three different objectives are being trained, we evaluate the models in either a task adaptation setting or a conditional autoencoder setting and plot them separately.
If there is no loss on a validation set (task adaptation), the model is still able to predict locations in the autoencoder setting.
However we see that the model overfits to the task conditioning examples and does not generalize to new views.
On the other hand, if we use only the validation set loss, the model is still able to learn, but training is slower and less stable.
Our hypothesis is that the autoencoder loss helps stabilize early training, but fails to force the model to decouple point \textit{identity} and \textit{position}.

\vspace{-0.1cm}

\subsection{Off-Surface Tracking}
\label{sec:off_surface_tracking}

For robotic grasping or insertion applications, the points of interest for control often do not lie on the surface of the object \cite{vecerik2020s3k},
but instead within the object (\eg the centre of mass).
Furthermore the ability to train on non-surface points is critical for using datasets without surface correspondences, which are not easily available in the real world (which we exploit in \sref{sec:robot_experiments}).
In this section we explore the model under such conditions.

\begin{figure*}
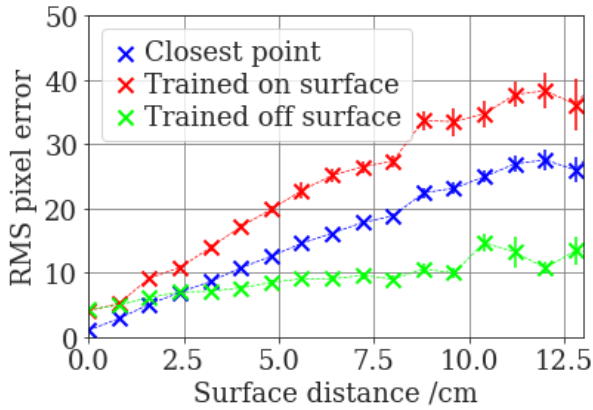

\centering
\begin{subfigure}{0.45\textwidth}
  \centering
\includegraphics[width=\textwidth]{plots/distance_error.pdf}
\end{subfigure}%
\begin{subfigure}{0.45\textwidth}
  \centering
\includegraphics[width=0.9\textwidth]{plots/offsurface_visualisation.pdf}
\end{subfigure}
    \caption{
       Evaluation of off-surface point prediction on withheld test shoes.
       Left: When training on off-surface points, TACK outperforms a model only trained on-surface points and a closest surface point oracle.
       Right: Visualization of methods given example points 10cm the off-surface.
       3 annotated examples were provided for each of the methods.
       The red and green overlays are the heatmap outputs, blue is the closest point oracle, and the white cross is the tracked location.
    }
  \label{fig:distance_error}
  \vspace{-0.5cm}
\end{figure*}

To test this property, we trained TACK on 3D points sampled from a distribution ${x'}^\Tau = \mathcal{N}(x^\Tau, \sigma^2)$, where $x^\Tau$ is a point on the object surface.
In this experiment we use $\sigma = 0.05$cm for all three dimensions.
For evaluation, we sampled 3000 points around each of the test shoes according to the same distribution.
For each point we computed their distance to the surface and predicted their location from a novel view.
Note that the points could be inside the shoe, but for the sake of simplicity \fref{fig:distance_error} considers the magnitude of surface distance and not the direction.

To test if the model can reason effectively about the off-surface points, we compared it to a closest point tracker oracle.
Given an off-surface point at test time, this tracker picks the closest point on the surface of the object, and uses knowledge about the camera views and the object geometry to project it into the new camera view. We then measure the RMS pixel error between the off-surface point and the closest on-surface point.
This gives us a lower bound on the error achievable by models which can only track points on object surfaces, such as any dense descriptor model.

\fref{fig:distance_error} compares the off-surface point model, the closest point oracle, and a model trained only with on-surface points. For points close to the surface (\eg $<$2cm) there is little difference between the three models.
However for larger distances (\eg $>$8cm), the ability of the off-surface model to reason about relative locations to the object allows it to outperform both baselines.

\subsection{Within-Class Generalization}
\label{sec:generalization}

For robotics applications, the ability to annotate a point on a single object and identify the same point on novel instances from the same class is a desirable property.
A naive way to train such a detector would require labels which correlate the same semantic points across instances of objects.
However, TACK learns a spatial embedding space that is reused across 56 different object instances which becomes cross-instance consistent in an unsupervised way.

To demonstrate this we chose 4 shoes unseen during the training and hand-picked a position and orientation for each, for ease of visual comparison.
Next we sampled 14 points on the surface of the first shoe and grouped them into 7 (\textit{start, end}) pairs.
Interpolation between points in embedding space and detecting their locations creates a curve on each of the shoes in image space, shown in \fref{fig:embedding_interpolation}.
The interpolated curves are qualitatively similar across all of the shoes despite not being explicitly trained for consistency.

\begin{figure}[]
  \centering
  \vspace{0.1cm}
\includegraphics[width=0.5\textwidth]{plots/embedding_interpolation.pdf}
    \caption{
        We pick 7 embedding pairs based on random points on the first shoe and interpolate linearly between pairs in embedding space.
        We detect the interpolated embeddings
        and draw the detected points in different colors for each original pair.
        The same embeddings correspond to similar locations on each shoe, showing that TACK has learned a consistent mapping across 
        instances.
    }
  \label{fig:embedding_interpolation}
  \vspace{-0.5cm}
\end{figure}

\section{Comparisons to other methods}

We consider the three approaches most closely related to TACK: S3K\cite{vecerik2020s3k}, kPAM\cite{Manuelli2019kPAMKA}, and Dense Object Nets\cite{florencemanuelli2018dense}.
Both S3K and kPAM focus on the sparse case in which the model has an explicit output dimension for each point to be detected.
Specific points are obtained by indexing this output, thus the keypoint identity can be viewed as a 1-hot vector representation.
By contrast, Dense Object Nets generate a single pixel-wise feature embedding, and obtain keypoint heatmaps via dot-product with a query vector followed by a softmax operation.
TACK can be seen as a hybrid between these approaches as it is trained explicitly for keypoint detection as in kPAM or S3K, but densely for every point, like Dense Object Nets.

We use fully-supervised S3K as the sparse-keypoint baseline in this work, as it shares the most with the setting we consider here.\footnote{
Although  kPAM assumes full 3D keypoint supervision similar to TACK, it also learns depth in addition to 2D coordinates, rather than triangulating from multiple cameras.}
Based on published results \cite{vecerik2020s3k} we expect S3K's self-supervision mechanism to require at least 10 3D labels (20 image annotations) to converge.
Beyond this the detection accuracy is equivalent to fully-supervised models and we can approximate its performance by using a fully supervised dataset - similar to kPAM.
This is significantly more than TACK which uses only up to 3 annotations. 

To generate a dataset we subsample object meshes using farthest point sampling, and project these 3D points to image coordinates from multiple views.
Importantly, we do not assume cross-instance labels, \eg there is no label indicating that the tip of shoe A is equivalent to the tip of shoe B.
This is a weaker form of supervision than kPAM, and forces the model to learn embeddings purely based on visual similarity.

Since our datasets do not have cross-instance labels, we focus on two scenarios: learning about a single instance, or learning about 8 instances where the labels are not semantically consistent.
In all cases we evaluate the RMS pixel-error on the training keypoints from novel views.

To summarize we compared four approaches:
\begin{enumerate}
    \item \emph{Supervised S3K.}
Uses the same model as TACK decoder, but outputs a heatmap for each keypoint instead of a single one.
Trained with a heatmap cross-entropy loss similar to TACK, but can only be evaluated at the specific points on which it was trained.
    \item \emph{Canonical TACK.}
Trained on random points as in \sref{sec:dataset}, but evaluated on the same subset of points as other models.
    \item \emph{Subset TACK.}
Same architecture as \emph{Canonical TACK}, but only trained  on the specific training points seen by S3K.
    \item \emph{Dense Objects Nets (DON).} Trained on dense cross-image correspondences.
\end{enumerate}

\begin{figure*}[h]
  \centering
  \vspace{-0.1cm}
\includegraphics[width=0.9\textwidth]{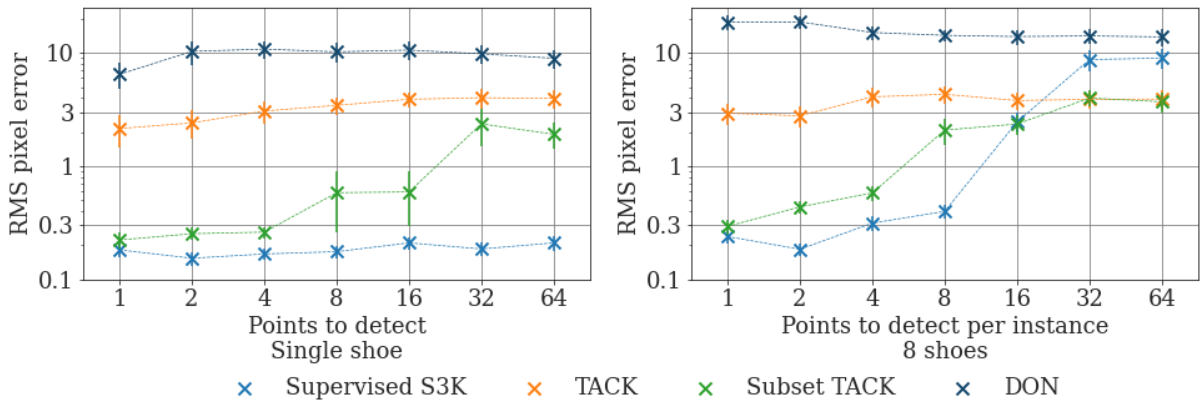}
\vspace{-0.3cm}
    \caption{
    Comparison of TACK to baselines.
    The left plot shows detection accuracy on a single shoe dataset.
    The right plot shows the same on an 8 shoe dataset for a case where our labels are not geometrically consistent across instances.
    }
  \label{fig:kpam_comparison}
\vspace{-0.6cm}
\end{figure*}

\subsection{Results}
In the left plot of \fref{fig:kpam_comparison} we see that the fully supervised baseline, which acts as an upper performance bound, can make use of the identity labels to outperform TACK by about a factor of 10.
This is because TACK is a dense tracking model which infers the point's identity as well as its location.
If we compare it to a model with the same objective such as Dense Object Nets, TACK outperforms it by a factor of about 3.
For small numbers of points approach 3) leads to a higher accuracy than TACK, because the identification task is simpler as TACK's latent space becomes more discrete.
Increasing the number of points to track leads to a continuous distribution associated with a lower precision.

The RMS pixel errors for 64 keypoints and 8 objects, as shown on the right side of \fref{fig:kpam_comparison}, correspond to 9.0 for the fully supervised model, 3.9 for full TACK, 3.7 for TACK trained on a subset of points, and 13.9 for Dense Object Nets.
On the right hand side we can see that TACK performs almost as well in the cross-instance generalization setting as it does on a single instance.
However as the number of points increases, the fully supervised baseline's performance deteriorates rapidly.
This happens because the supervised baseline has a different combination of labels for each instance, and therefore it cannot learn to generalize across instances (\ie the first keypoint might be on the tip of shoe 1, but on the heel of shoe 2).
We can see that in this case the subset TACK outperforms the S3K baseline.
The reason for this is that when keypoints are not matched across different instances, S3K has to implicitly learn about these permutations.
Subset TACK on the other hand is allowed to generalize across the instances as it infers them itself from the conditioning images and is not exposed to the underlying cross instance inconsistency of the dataset.

\section{Robot Experiments}
\label{sec:robot_experiments}

To demonstrate our method's real-world applicability, we used it as a perception module to solve a grasping task with a real robot.
We use a both a real and simulated datasets to achieve this.

\subsection{Real World Dataset}
\label{sec:real_world_dataset}

Without assuming access to a ground-truth depth estimation we cannot directly sample object surface points as we did in simulation.
However, one of the main advantages of TACK over other dense descriptor techniques \cite{florencemanuelli2018dense} is the ability to represent 3D keypoints \textit{relative} to an object, which allows us to provide weaker supervision on the real-world dataset.
Our hypothesis was that training on off-surface points was sufficient to adapt the model on real-world data, and allow the embeddings to transfer.

Specifically, we implemented a rejection sampling scheme which allowed us to sample 3D points that were near the object.
This was done by accepting points whose projected pixel corresponded to background.
This scheme required knowing the approximate 3D position of the object, and the background color (a segmentation would also suffice).
We collected 5000 samples for 3 different shoes in the real-world dataset.
Since our simulation experiments used a black background these models would not transfer well, so we performed a manual foreground-background subtraction and segmented out the gripper, which is fixed in the camera-frame.
We then used the segmented scene as a background for the simulated dataset, which we first overlaid with the rendered object, and finally with the gripper to generate realistic occlusions.
We combined the simulated and real datasets by using 30 augmented simulation images for every 2 real images per training batch.
Note that this strategy for generating a real dataset is only applicable because TACK is capable of learning about off-surface points, as suggested in \sref{sec:off_surface_tracking}.

\subsection{Grasp Controller}

Performing a grasp task requires knowing the grasp point from a correct direction.
After using TACK to detect grasp points, we can script robot motions to achieve a successful grasp.
We used a single shoe to annotate the heel and centre of the shoe.
Using this, we can detect these points on other shoes.
We used these two points to compute references for a grasp controller, which uses a series of p-controllers to command the robot's end effector to the desired position.
This worked on various shoes including ones which were not seen during training in either of the two datasets.

This result suggests an alternative path towards generalized pick and place: instead of attempting to train the full grasp policy a priori, we can instead train a perception module using TACK, and quickly specify the policy \emph{in situ}.
In advantage to being easier to train, this offers transparency to the end-user, and the flexibility to novel policies.

\subsection{Results}

The robot was able to pick, rotate, move, and place the shoe at different start and end poses. Different shoes were swapped into the scene and the robot was able to robustly grasp these shoes at the keypoints corresponding to the initial two grasp points.
In addition to the three shoes used in training, the robot can robustly grasp four withheld test shoes of varying shapes, sizes and colors.
The video of the robot grasping demo is available in the supplemental material and the project website.

\section{Conclusion}
\label{sec:conclusion}

We presented an approach which formulates 3D keypoint tracking as a combination of task adaptation and conditional autoencoder problems.
We showed how this formulation relates to other approaches such dense embedding learning, while presenting a unique set of advantages:
it can learn about novel locations from only a few examples ($<$3), it can reason about points off the surface, and generalizes across instances of the same class.
Finally we showed that TACK allows the use of real robot data for tracking when no explicit cross-view correspondences are available, \ie using calibrated RGB cameras with no depth information.
We demonstrated using TACK in a robot controller which can grasp any shoe using only a few annotated grasp points.

As future work, TACK could be incorporated into a reinforcement learning agent.
Here it could be used as a representational bottleneck either for fast visual learning or Sim2Real policy transfer.

\bibliographystyle{IEEEtran}
\bibliography{main}

\appendicess

\section{Network diagrams}
\begin{figure*}[h]
  \centering
\includegraphics[width=0.9\textwidth]{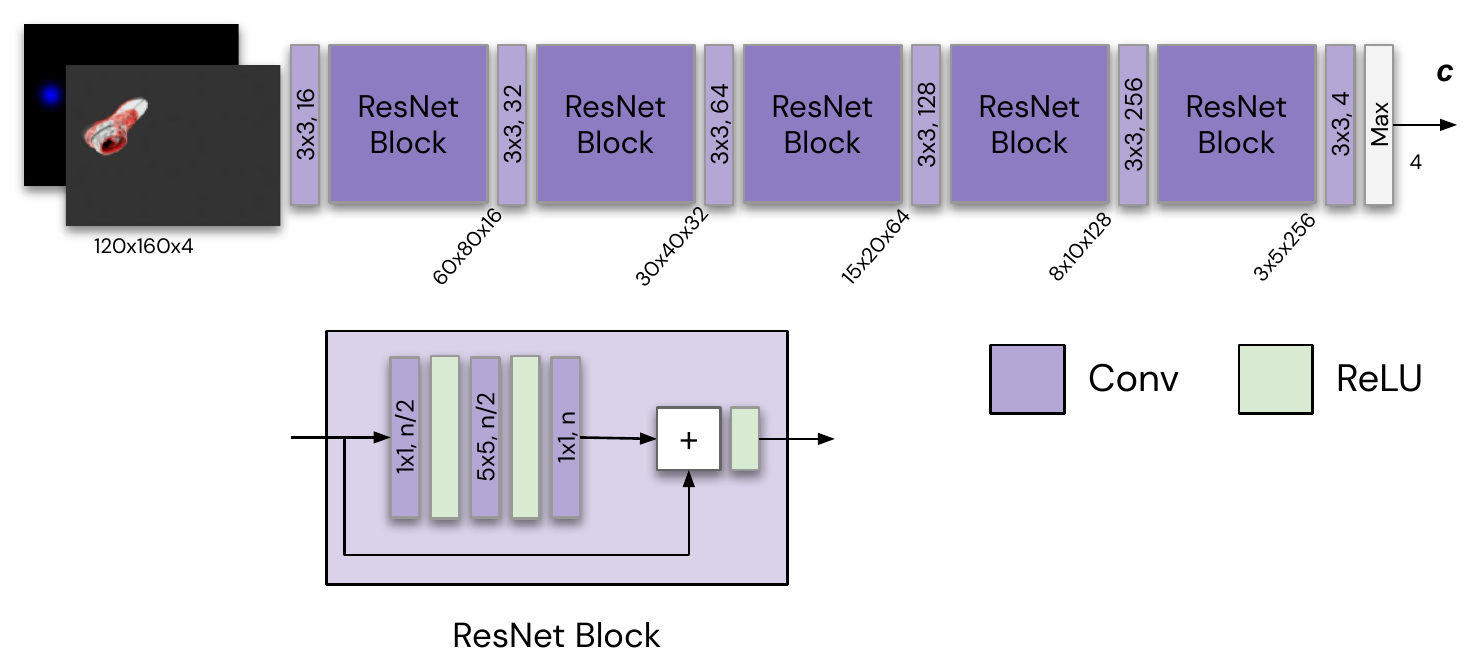}
    \caption{
        Encoder network diagram. The image and target are concatenated in the channel dimension and passed to 5 layers of convolutions and ResNet blocks. The final embedding has size 4.
    }
  \label{fig:encoder_diagram}
  \vspace{-0.5cm}
\end{figure*}

\begin{figure*}[]
  \centering
\includegraphics[width=0.9\textwidth]{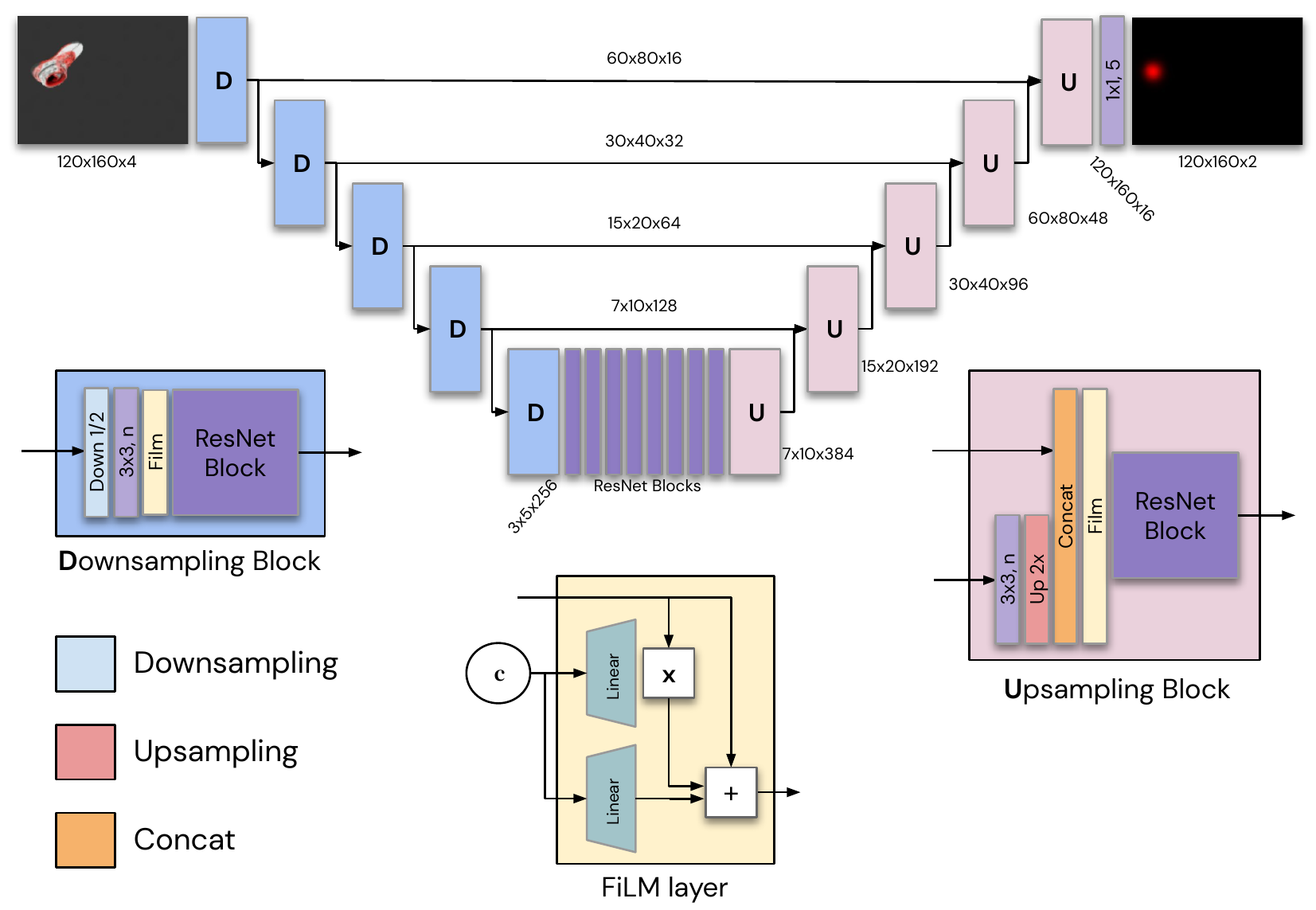}
    \caption{
        UNet-inspired decoder network diagram, composed of a series of 5 downsampling blocks and 5 upsampling blocks that receive the downsampled image from the corresponding layer as input.
    }
  \label{fig:decoder_diagram}
  \vspace{-0.5cm}
\end{figure*}

In \fref{fig:encoder_diagram} and \fref{fig:decoder_diagram} we present detailed diagrams of the network architectures described in \sref{sec:network_architecture}.

\section{Number of Annotations}
The annotations are the only source of information about the point identity.
Therefore more annotations should increase the precision of the model.
Since the embeddings are aggregated by a simple average we can evaluate our model with various numbers of annotations provided. 

To investigate this, we designed an experiment to explore detection accuracy vs. the number of annotations used to query the model on test time.
Note that the model was still trained only with 3 conditioning images.
As discussed above, the model is always trained with three annotations, but can be queried with as few as one annotation to determine the keypoint to be tracked.
However, increasing the number of annotations and averaging the embeddings can improve the accuracy of the detection.

\begin{figure}[b]
  \centering
\includegraphics[width=0.4\textwidth]{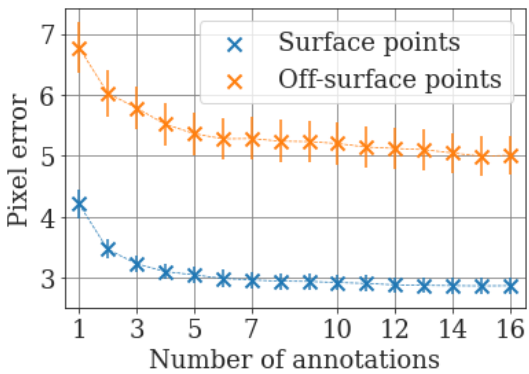}
    \caption{
        Keypoint accuracy as a function of the number of training annotations. The model was always trained on 3 annotated images, but test-time accuracy increases as we provide more annotations for a given point. 
    }
  \label{fig:number_of_annotations}
  \vspace{-0.5cm}
\end{figure}

To conduct this analysis, we sampled 800 meta-batches containing 16 training and 4 test images for a given point.
We then computed embeddings using between 1 and 16 of the training images, and evaluated the mean tracking accuracy on the remaining 4 test images.
\fref{fig:number_of_annotations} shows that providing more annotations monotonically improved the accuracy of the model.
However even with just 1 annotation we were able to track the desired point, as demonstrated in \sref{sec:robot_experiments}.
This analysis was performed separately for surface and off-surface points because of the significant difference in accuracy between these settings (presumably, off-surface points are harder to localize because there is their relative position is more ambiguous than surface points).
A model trained on off-surface samples is used for this experiment.

\section{Choosing the Embedding Size}
\label{sec:embedding_size}

\begin{figure}[h]
  \centering
\includegraphics[width=0.4\textwidth]{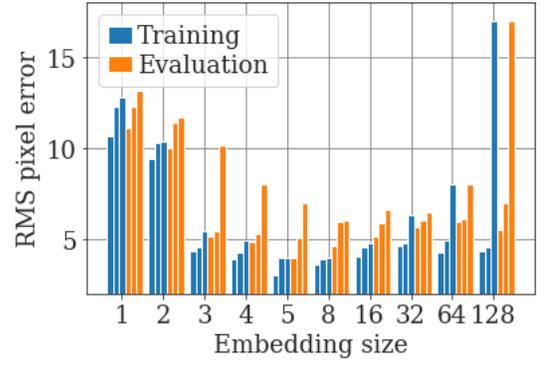}
    \caption{Tracking accuracy as a function of the embedding size. If the embedding is too small, it cannot contain enough information to accurately track keypoints. However, too large of an embedding results in overfitting and more variance across seeds.
    }
  \label{fig:embedding_size}
\end{figure}

The embedding $c^\Tau$ carries information about the identity of a point on or relative to the object so its dimensionality is an important hyperparameter.
Since the keypoint can be off the surface, the embedding must have sufficient capacity to represent a 3D offset, in addition to the location(s) on the object that the keypoint is relative to.
To identify the required capacity of $c^\Tau$ we performed a sweep over the dimensionality $K$ of the embedding using the simulated shoe dataset.

For each $K$ ranging from 1 to 128 we performed 3 training runs and measured the final test set performance.
\fref{fig:embedding_size} shows both the training and validation performance of each independent run (to illustrate the variance).
As expected, $K<3$ was insufficient to capture the 3D offset of the tracked points.
However we found that $K>3$ can stabilize training, as seen by the lower mean and variance for embedding sizes larger than 3, up to a maximum of around 32.
Beyond $K=32$ we observed over-fitting to the training set.
Following this analysis we used $K=4$ as our embedding size in all other experiments.

\section{Conditioning the Decoder}

The Decoder's task is to consume an image $\img$ and task embedding $c^\Tau$ and detect the required point $p^\Tau$ in the image:
\begin{equation}
    Dec(c^\Tau, \img) \rightarrow p^\Tau
\end{equation}
We have to carefully choose an architecture as the image is represented by a 2D array of RGB values while the embedding is represented by a single low dimensional vector.
Therefore we need an architecture which can use both information sources efficiently.
In this section we explore options in this choice.

In TACK we represent the output detection as a heatmap, i.e. $p^\Tau \in \mathbb{R}^{(H,W,1)}$, which motivates a fully-convolutional model.
As discussed in \sref{sec:network_architecture}, our decoder is a ResNet \cite{He2015Residual}, whereas our embedding $c^\Tau \in \mathbb{R}^K$ is a vector.
This precludes using standard vector-concatenation at a bottleneck to condition the decoder.

\begin{figure}[h]
  \centering
\includegraphics[width=0.4\textwidth]{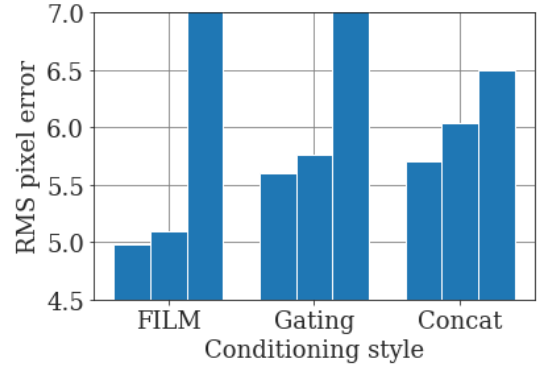}
    \caption{
    Comparing evaluation error across the decoder conditioning functions for 3 seeds (sorted by error). FiLM has the best average performance across seeds.
    }
  \label{fig:conditioning_type}
  \vspace{-0.5cm}
\end{figure}

We explored three strategies for conditioning the decoder such that it can effectively use information in the task embedding.
These approaches are summarized below, and compared in \fref{fig:conditioning_type}.

\begin{enumerate}
    \item \textbf{Concat}: tile embedding $c^\Tau \in \mathbb{R}^K$ across the image dimensions to produce a feature map of shape $(H, W, K)$, and concatenate with $\img$ along the channel dimension. In this approach we only condition $Dec$ at the input.
    \item \textbf{Gate}: for each selected layer $l$, project the embedding to the required feature-dimension $f_l$ using a learned linear-layer $Linear_l(c^\Tau \in \mathbb{R}^K) \rightarrow c_l \in \mathbb{R}^{f_l}$, apply a sigmoid, and multiply the feature-maps along the channel-dimension (broadcasting across the feature dimensions (H, W)). \ie $x = x * sigmoid(Linear_l(c^\Tau)))$. This operation is applied within all upsampling and downsampling blocks after the up/down-sampling + $3x3$ conv, but before the residual block.
    \item \textbf{FiLM}: identical to \textbf{Gate}, except that we apply both shift and scale operations, rather than simply scale (and omit the sigmoid).  \ie $x = x * (mlp_l^{\sigma}(c^\Tau) + mlp_l^{\mu}(c^\Tau)$. 
\end{enumerate}

Although originally designed for visual question-answering \cite{Perez2018film}, \fref{fig:conditioning_type} shows that the FiLM approach yielded the best performance, and was used in all other experiments.

\section{Determining 3D Points}

Despite implicitly encoding 3D information, the output of the TACK decoder is still a 2D quantity.
However, given detections from multiple views we can compute the 3D point by solving least-squares problem that minimizes the distance between rays from multiple views through a single keypoint.
Defining the 3D position of camera $i$ as $\mathbf{a}_i$ and the normalized direction of the ray from camera $i$ as $\mathbf{\hat{d}}_{i}$, the following equation yields the estimated 3D keypoint location:

\begin{equation}
    \widetilde{\mathbf{x}}_k = \left(\sum_i^L \mathbf{I} - \mathbf{\hat{d}}_{i} \mathbf{\hat{d}}_{i}^\intercal \right )^{-1} \left ( (I - \mathbf{\hat{d}}_{i} \mathbf{\hat{d}}_{i}^\intercal) \mathbf{a}_i \right ) 
    \label{eq:unsup_solution}
\end{equation}

$\mathbf{d}_{i}$ is obtained by taking the spatial soft-argmax 
of the heatmap for some task $\Tau$ in camera $i$ to obtain the 2D keypoint, and then projecting it to a world-frame ray using the inverse camera matrix:
\begin{align}
    \mathcal{P} = softmax(p_i^\Tau) &&
    \mathbf{d}_{i} = {}^w T_i \  K_i^{-1} \left[ \mathbb{E}_\mathcal{P}(x), \mathbb{E}_\mathcal{P}(y), 1 \right]^T
    \label{eq:camera_rays}
\end{align}
where $K_i$ is the intrinsics matrix for camera $i$, ${}^w T_i$ is the homogeneous transform from camera $i$ to the world-frame, and $\mathbb{E}_\mathcal{P}(x), \mathbb{E}_\mathcal{P}(y)$ denotes the spatial soft-argmax over the image dimensions \cite{vecerik2020s3k}.
Given \eref{eq:camera_rays}, $\mathbf{\hat{d}_{i}}$ is obtained by normalization.

An issue with the naive least-squares formulation above is that it is not robust to outliers, which occurs frequently when one of the cameras becomes occluded or is queried out-of-distribution.
Prior work \cite{vecerik2020s3k} proposed a weighted-least-squares solution, in which the heatmap variance was used as a weight on the R.H.S. of \eref{eq:unsup_solution}.
In our early experiments we found that this solution failed to handle bi-modal heatmaps, in which the mean is invalid but one of the modes is likely correct according to other cameras, as well as diffuse heatmaps in which the variance is high \eg due to occlusion, but still exerted a non-zero influence on the 3D solution.

In this work we adopted the more explicit approach of maximizing a detection score over all possible combinations of cameras.
This approach provides a hard selection of which cameras to include in the estimate, while still allowing all cameras to vote on the best keypoint estimate.
Specifically, for each subset $\bar{C} \subseteq C$ containing $2$ or more cameras ($11$ possible for $4$ cameras) we compute the 3D point $\widetilde{x}_{\bar{C}}$ using \eref{eq:unsup_solution}, reproject back to \textit{all} cameras, and evaluate the heatmaps to obtain a detection score:
\begin{equation}
    \mathbb{S}(\widetilde{x}_{\bar{C}}) = \sum_{i=1}^L exp(p^\Tau_i - max(p^\Tau_i))(K \widetilde{x}_{\bar{C}})
\end{equation}
If a point projects outside of the image we use 0 for that image but we still consider it as a valid candidate.
Note that this formulation is a heuristic rather than a proper probabilistic treatment, however we found it to work well given that the model tends to be confident even when outputting wrong predictions.
The final point is selected from the subset with the maximal score.
We found this to be more stable and accurate than the weighted-least-squares solution, and was used to produce robot demo in \sref{sec:robot_experiments}.

\section{Training on a single object}

\begin{figure}[h]
  \centering
\includegraphics[width=0.5\textwidth]{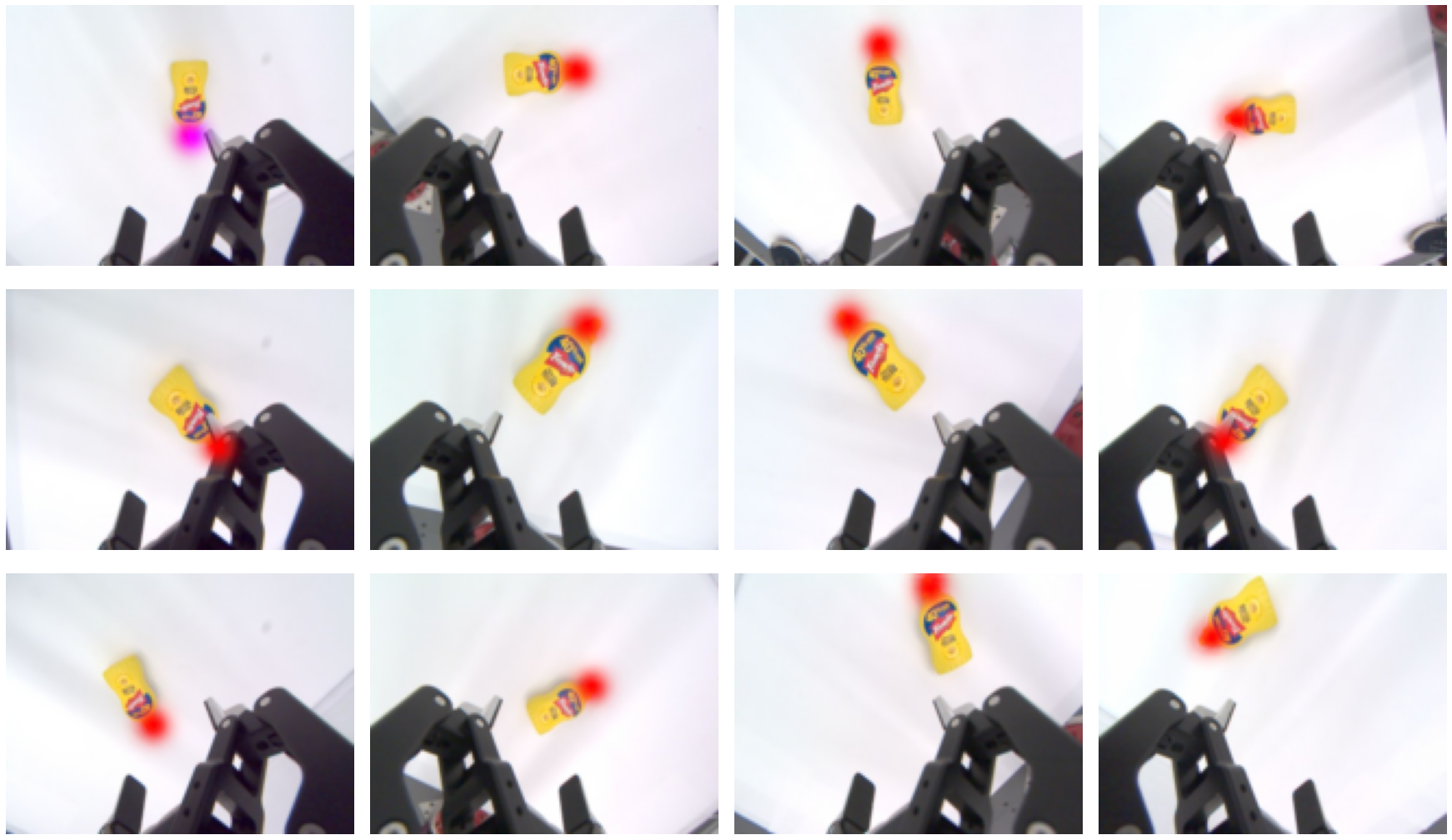}
    \caption{
    TACK detection with a model trained on the YCB mustard object.
    A single annotation was provided for the image in the top-left.
    Each column corresponds to a different camera and each row to a different timestep.
    The predicted heatmap from the model can be seen as red.
    }
  \label{fig:mustard_times}
\end{figure}

In all simulated experiments within this paper, we evaluated TACK in the context of in-class generalization.
However this setting is not a restriction of the model, and TACK can be trained on a single object instance.
To show this, we trained a model on a YCB mustard\cite{ycb2015Calli} with the method descrbed in \sref{sec:real_world_dataset}.
The results demonstrate this model's detection capability on a well known object from another class.
In \fref{fig:mustard_times} we visualize detections based on a single annotation.
We see that TACK is in most cases able to detect the rough position of the keypoint, but fails for large occlusions which significantly mask the shape.
See the supplementary material for a video illustrating this, as well as a video of detections while handling the object.

\subsection{Visualising Embeddings}
\label{sec:visualize_embeddings}

Another qualitative analysis which is useful for understanding model generalization is to directly visualize the embeddings as a color-space.
This is possible for dense descriptor models because they output an embedding for every pixel.
If the embedding space has size 3, we can normalize the embedding output and visualize it as a 3-channel RGB image.

\begin{figure}[]
  \centering
\includegraphics[width=0.45\textwidth]{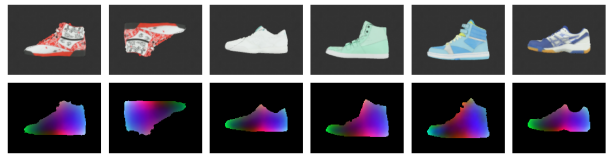}
    \caption{
        We treat every pixel as a tracking point and visualize the corresponding embedding.
        We can see a strong visual correspondence across the different shoe instances and orientations.
    }
  \label{fig:visualize_embeddings}
  \vspace{-0.5cm}
\end{figure}

For TACK there is no 1-to-1 correspondence between pixels and the embedding space.
However, we can select each point in the field of view in the camera as a target, encode the target, and convert the resulting matrix of embeddings to an image.
Since we use a 4-dimensional embedding space we ignore the last dimension, normalize the remaining 3, and interpret them as RGB channels.
As shown in \fref{fig:visualize_embeddings}, the visualized embeddings are semantically similar parts of the shoe.

\section{Other object occlusions}

\begin{figure}[]
  \centering
\includegraphics[width=0.4\textwidth]{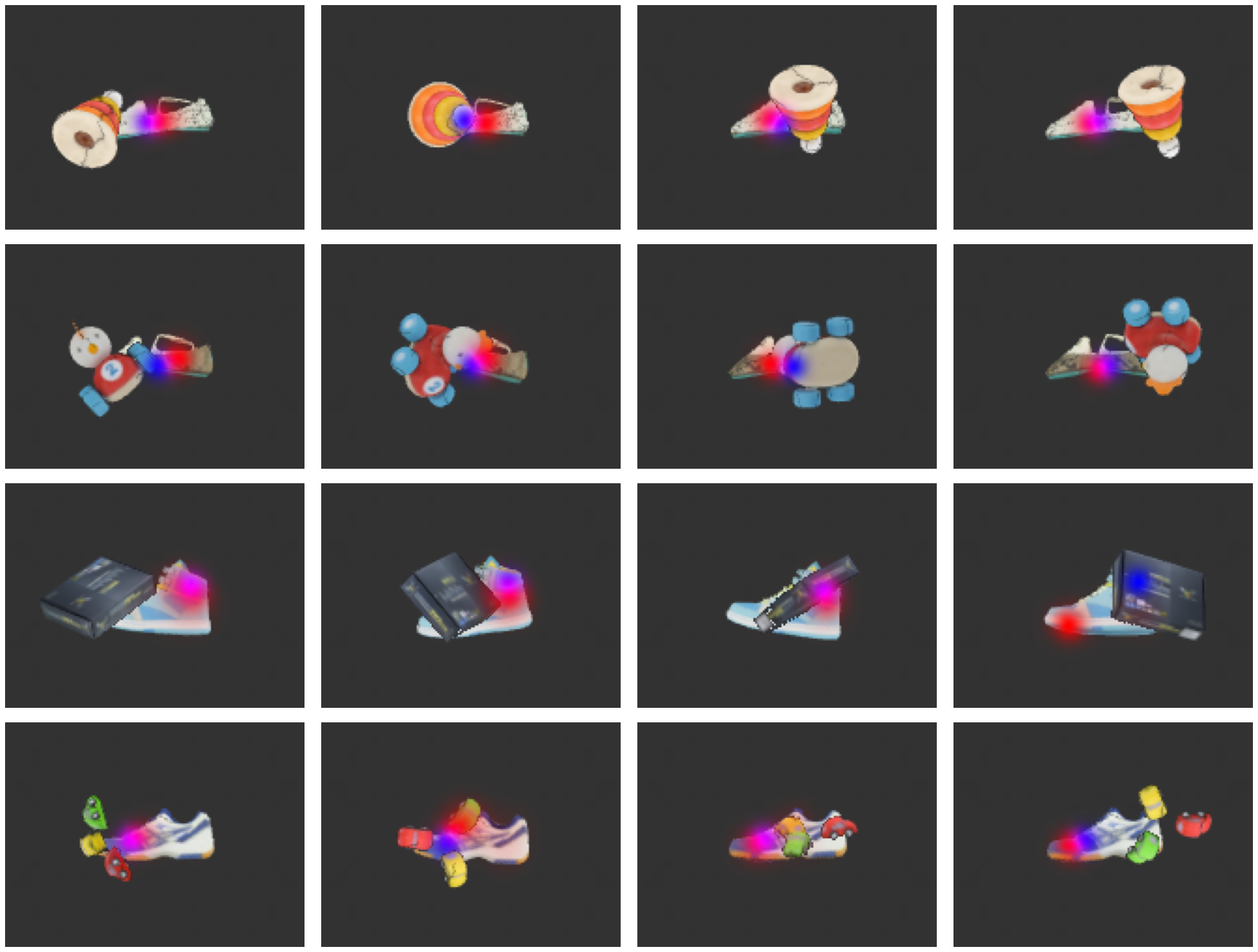}
    \caption{
    Using TACK to predict keypoints through occlusions by other objects.
    }
  \label{fig:occlusions}
\end{figure}

So far, we have shown that TACK can learn to deal with self-occlusions.
However the same properties allow it to deal with occlusions from other objects.
To demonstrate this, we trained a TACK model on a dataset which in addition to 1 shoe included 2 other distractor objects with random poses.
For qualitative evaluation, we acquired the keypoint embedding based on a single an image where the shoe was visible, then performed multiple detections as a distractor object moved in front of it.
Keeping track of the point is a challenging task. 
In \fref{fig:occlusions} we show samples from these detections.

\section{Gradient analysis}
\begin{figure}[]
  \centering
\includegraphics[width=0.4\textwidth]{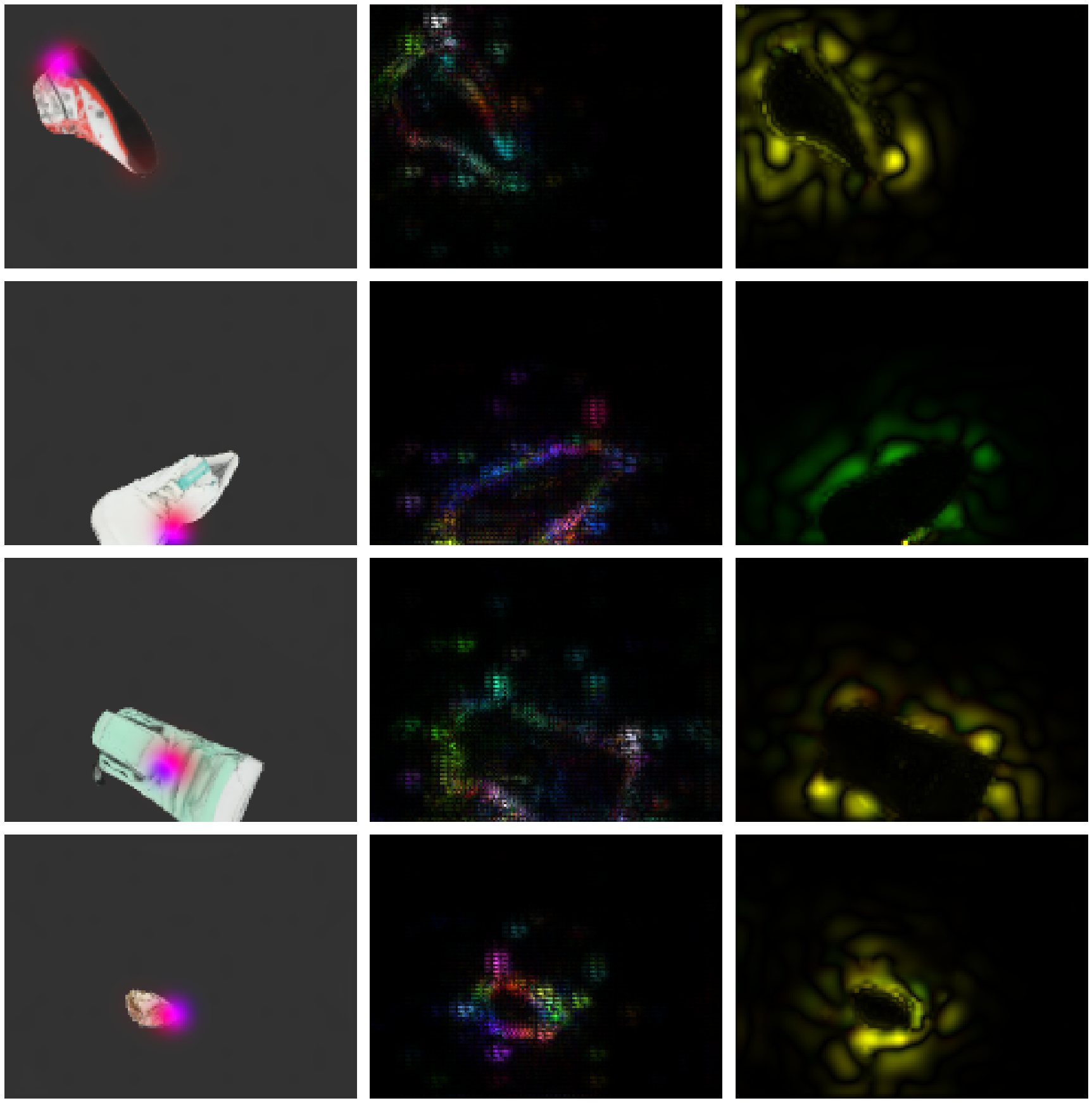}
    \caption{
    Saliency analysis of the TACK model.
    We used a single image autoencoding setup for these experiments.
    The first column shows the image, target (blue) and the model prediction(red).
    The second column shows the magnitude of the endcoder gradient for the first 3 components of the embedding vector with respect to the input image.
    The last columns show the decoded position gradients.
    The x gradient is represented by the red and y component by the green channel. 
    }
  \label{fig:grad_cam}
  \vspace{-0.5cm}
\end{figure}
In \sref{sec:visualize_embeddings} we provided experiments which relate the structure of the embedding space to the image location.
For the sake of interpretability, we are also interested in how the outputs of neural networks depend on their input values, and what properties of the data the networks are sensitive to.
To explore this we chose saliency maps as in \cite{Saliency2014Simonyan14a} to explore both the encoder and decoder models.

Saliency maps and similar visualisation approaches were originally developed for classification models where we compute the gradient of a specific class logit with respect to the input image, but we slightly modify the formulation to visualize TACK's embedding space.
Our encoder network has 2 inputs (image and target annotation) and outputs an embedding vector (usually 4-dimensional).
For our encoder we decided to explore the gradient of the first 3 elements of the embedding vector with respect to each pixel of the input image.
These 3 gradients can be then mapped onto the 3 RGB image channels.
Our decoder has also 2 inputs (image and embedding), but it outputs a full per-pixel heatmap.
Thus there is no natural low-dimensional output to use for gradients.
To resolve this, we reduce the output to the expected x and y values using the spatial softmax as described in \sref{sec:experiments}.
This allows us to compute the gradient of these expectations with respect to the input image.
We visualise the normalised magnitude of the x, y gradients as RG channels respectively.

We present these heatmaps on a simulated dataset in \fref{fig:grad_cam}.
We can see that both the decoder and the encoder focus mainly on the outline of the shoe, suggesting that these networks naturally focus more on the overall shape of specific locations, rather than the texture.
In the supplementary material we also show a video version of this analysis for a real dataset.
However for a real dataset we only have a single annotation, thus we cannot construct an appropriate input for the encoder for each frame.
Therefore we only focus on the decoder gradients.
Compared to the simulated dataset they are more focused on the shoe itself rather then just the outline.
This is likely because the model needs to distinguish the shoe from other objects such as the gripper or struts.

\section{Robot setup}

We used a Sawyer robot with a 10Hz proportional controller, a 2f-85 Robotiq gripper with 4 RGB Basler daA1280-54ucm cameras using F2.8 f2.95mm lenses.
We undistorted the 640x480 images and down-sampled to 1/4 resolution before passing them to the model.

\section{Hyperparameters}

We used a decaying learning rate which starts at $1e-4$ and decays to $1e-5$ over training.
The batch size is 32.
The task adaptation $\mathcal{L}_{adapt}$ and autoencoder $\mathcal{L}_{auto-enc}$ losses were summed together using equal weighs.
No explicit network regularization such as L1 or L2 was used.
The segmentation loss used a per-pixel cross entropy loss and was weighted with other losses with a weight of 0.1.
Disabling this loss did not significantly effect the performance on simulated datasets.
The $\sigma$ for our target images $t_{ij}$ was 5.

\section{Examples from Simulated Dataset}

\begin{figure*}[h]
  \centering
\includegraphics[width=0.8\textwidth]{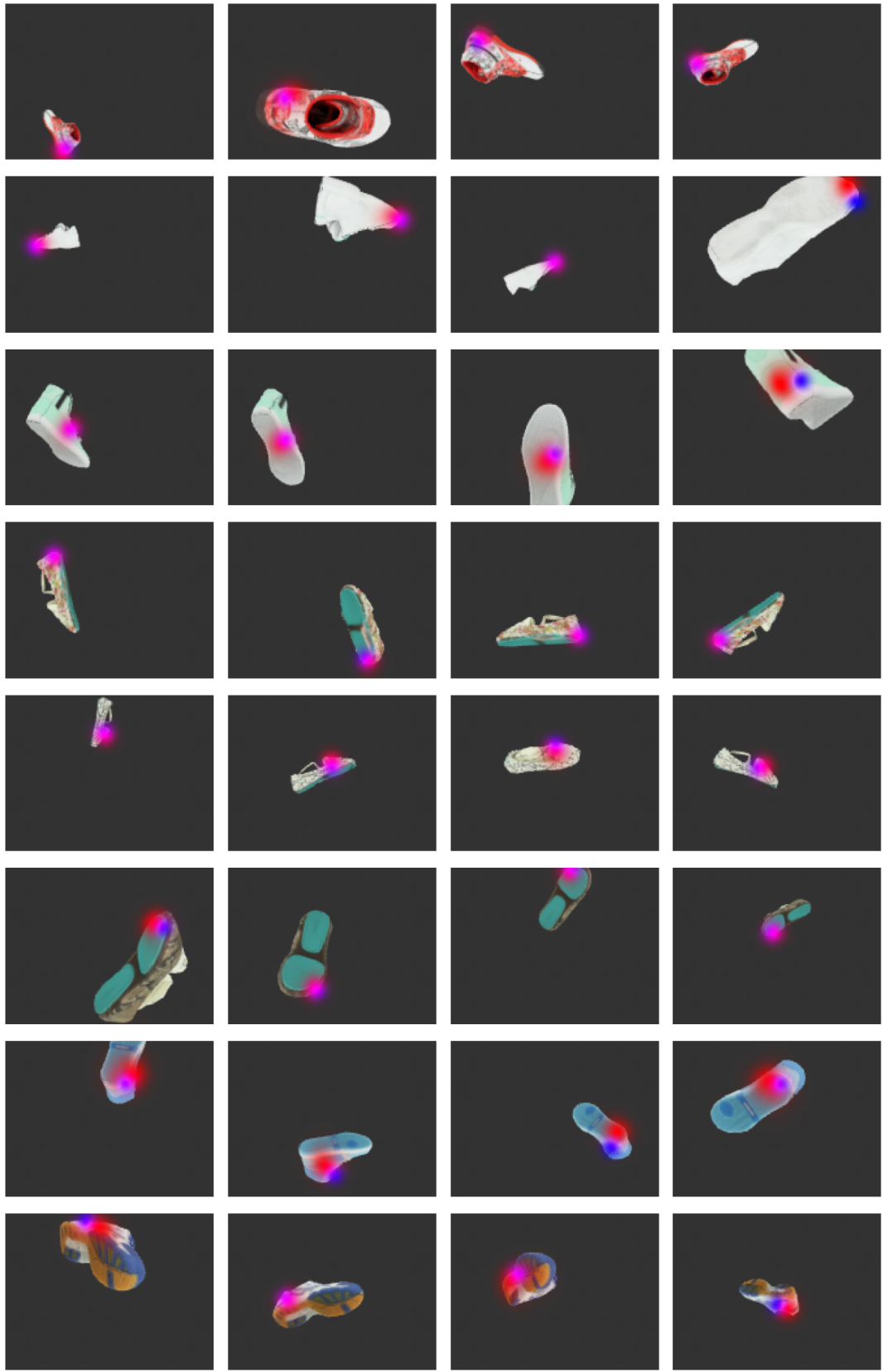}
    \caption{
        An example meta-batch for each shoe in the evaluation dataset.
        The images in the first 3 columns are used to determine point identity.
        Fourth column contains the validation sample.
        Red and blue overlays are the model prediction and target respectively.
    }
  \label{fig:dataset_examples}
\end{figure*}

In \fref{fig:dataset_examples} we show examples from the evaluation dataset and example model's detections.

\end{document}